\documentclass{ceurart}

\usepackage{algorithm}
\usepackage{algpseudocode}
\usepackage{booktabs}

\begin{document}

% \title[An Agentic Framework for Experience-Guided Strategy Refinement in E-Commerce Recommendation]{From Bad Cases to Better Strategies: An Agentic Framework for Self-Refining Recommendation Strategies in E-Commerce}
% \title[-Agent: Experience-Driven Self-Refinement Agent Framework]{SR-Agent: An Experience-Driven Agentic Framework for Post-Ranking Strategies Self-Refinement in E-Commerce Recommendation}

\copyrightyear{2026}
\copyrightclause{Copyright for this paper by its authors.
  Use permitted under Creative Commons License Attribution 4.0
  International (CC BY 4.0).}

% Replace this placeholder with the workshop's official name, date, and venue.
\conference{Recommender Systems Workshop, 2026}

\title{SR-Agent: An Experience-Driven Agentic Framework for Post-Ranking Strategy Refinement  in E-Commerce Recommendation}

% \title[SR-Agent: Strategy Refinement Agentic Framework]
% {From Bad Cases to Better Strategies: An Experience-Driven Agentic Framework for Post-Ranking Strategies Refinement in E-Commerce Recommendation}

% Replace the anonymous metadata below for the camera-ready version.
% \author[1]{Anonymous Author(s)}
% \address[1]{Anonymous Institution}
\author[1]{Hanchen Yang}
\fnmark[1]
\author[1]{Kaiwen Yang}
\fnmark[1]
\author[1]{Junpeng Zhuang}
\fnmark[1]
\author[1]{Yang He}
\author[1]{Keting Cen}
\author[1]{Bochao Liu}
\cormark[1]
\author[1]{Zhongbo Sun}
\cormark[1]
\author[1]{An Liu}
\author[1]{Zhongteng Han}
\author[1]{Chenyi Lei}
\address[1]{Kuaishou Technology, Beijing, China}
\fntext[1]{Equal contribution.}
\cortext[1]{Corresponding authors.}

\begin{abstract}
User experience is a first-class objective in industrial e-commerce recommender systems (RS). \emph{Post-ranking strategies}, which govern diversity, similarity, and exposure over a ranked list, are widely deployed in industrial RS for their simplicity and low serving cost. However, as the online recommendation environment evolves continuously, these statically configured strategies gradually become stale, thereby degrading the user experience. Refining them typically relies on manual inspection, diagnosis, and updates, making it slow, costly, and difficult to scale or reuse. Although recent LLM-based agents~(e.g., RecUserSim, SimUSER, and Self-EvolveRec) offer promising directions, none of them close the full loop of automated, self-evolving strategy refinement.
To bridge this gap, we introduce \textbf{SR-Agent}, which, to the best of our knowledge, is the first agentic framework deployed to refine post-ranking strategies in industrial RS.
SR-Agent unifies three components: (i) a \textbf{UserSim} agent that applies inspection skills to surface user-perceived bad cases; (ii) an \textbf{Analysis} agent that consolidates recurring bad cases into structured, reusable diagnoses; and (iii) a constrained \textbf{Strategy Refinement Harness} that maps diagnoses to typed and bounded actions, gated by a four-stage reward pipeline with reversible rollback. Deployed on the Kuaishou e-commerce platform, SR-Agent continuously runs this refinement loop and, in a one-month online A/B test, increases order volume by 0.71\%, browsing depth by 0.34\%, and clicked-category diversity by 0.48\%, while markedly shortening the refinement cycle and lowering operational cost.
\end{abstract}

\begin{keywords}
E-commerce recommendation \sep
multi-agent systems \sep
user experience \sep
large language models
\end{keywords}

\maketitle

\section{Introduction}

Improving user experience has always been a primary goal of e-commerce recommender systems (RS)~\cite{zhao2025llmagentrec_survey,adaji2017towards}. Beyond accurately identifying items that users are likely to click or purchase, an RS should organize its recommendations to provide a coherent and satisfying experience throughout the shopping journey~\cite{song2025user,pu2012userperspective}. However, optimizing pointwise relevance alone does not necessarily lead to a high-quality recommendation experience. For example, an RS may recommend multiple visually similar products because each item has a high predicted probability of being clicked or purchased. Despite their individual relevance, presenting these items together offers users limited alternatives and impedes efficient interest exploration. Such experience-level deficiencies can reduce user satisfaction and engagement, ultimately affecting purchase conversion and long-term retention~\cite{duricic2023beyondaccuracy,mcnee2006accurate}.

Over the past decades,  RS research has increasingly moved beyond predictive accuracy toward broader experience objectives such as diversity, novelty, and reduced repetition. Existing studies address these objectives from multiple angles~\cite{ziegler2005improving,kunaver2017diversity,lathia2010temporal,chen2018fastdpp}: \emph{re-ranking} methods (e.g., MMR, DPP) post-process candidate lists to trade off relevance against diversity or novelty; \emph{regularized training objectives} bake experience signals into the loss via multi-task learning; and \emph{reinforcement-learning-based policies} cast recommendation as sequential decision-making to optimize long-horizon rewards such as fatigue or return rate. Beyond these learning-based paradigms, a large family of rule-based \emph{ post-ranking strategies} has proven remarkably effective in practice, and has become a dominant mechanism in large-scale industrial RS. Their low serving cost, operational flexibility, interpretability, and seamless compatibility with existing pipelines make them particularly well suited to large-scale deployment.
Despite these advantages, two significant real-world challenges in industrial RS still need to be addressed.

% In past few decades, moving beyond accuracy to optimize the overall recommendation experience has become an important yet challenging problem in modern RS research. In practice, industrial RS address this challenge at multiple stages of the recommendation pipeline. They typically combine multi-objective ranking models~\cite{kunaver2017diversity,lathia2010temporal,chen2018fastdpp} with post-ranking strategies~\cite{wilhelm2018youtubedpp,zhao2025llmagentrec_survey} to balance item-level relevance with list- and session-level experience objectives.
% Among these mechanisms, \emph{post-ranking strategies} provide a practical way to adjust the final recommendation results without modifying the online ranking model. In this work, post-ranking strategies refer to the rules and parameterized controls that govern diversity, similarity, and exposure after ranking and before display.
% Typical strategies include rule-based quota controls over categories, shops, brands, or price range, etc; similarity-based suppression using visual or semantic embeddings; and diversification methods such as Maximum Marginal Relevance (MMR) and Determinantal Point Process (DPP)-based re-ranking, which reorder candidates by balancing relevance and diversity~\cite{ziegler2005improving,kunaver2017diversity}.
% These strategies are attractive because they are lightweight, interpretable, and efficient enough for online serving. Despite their practical importance, two real-world challenges remain to be addressed.

\begin{figure}[t]
  \centering
  \includegraphics[width=0.82\linewidth]{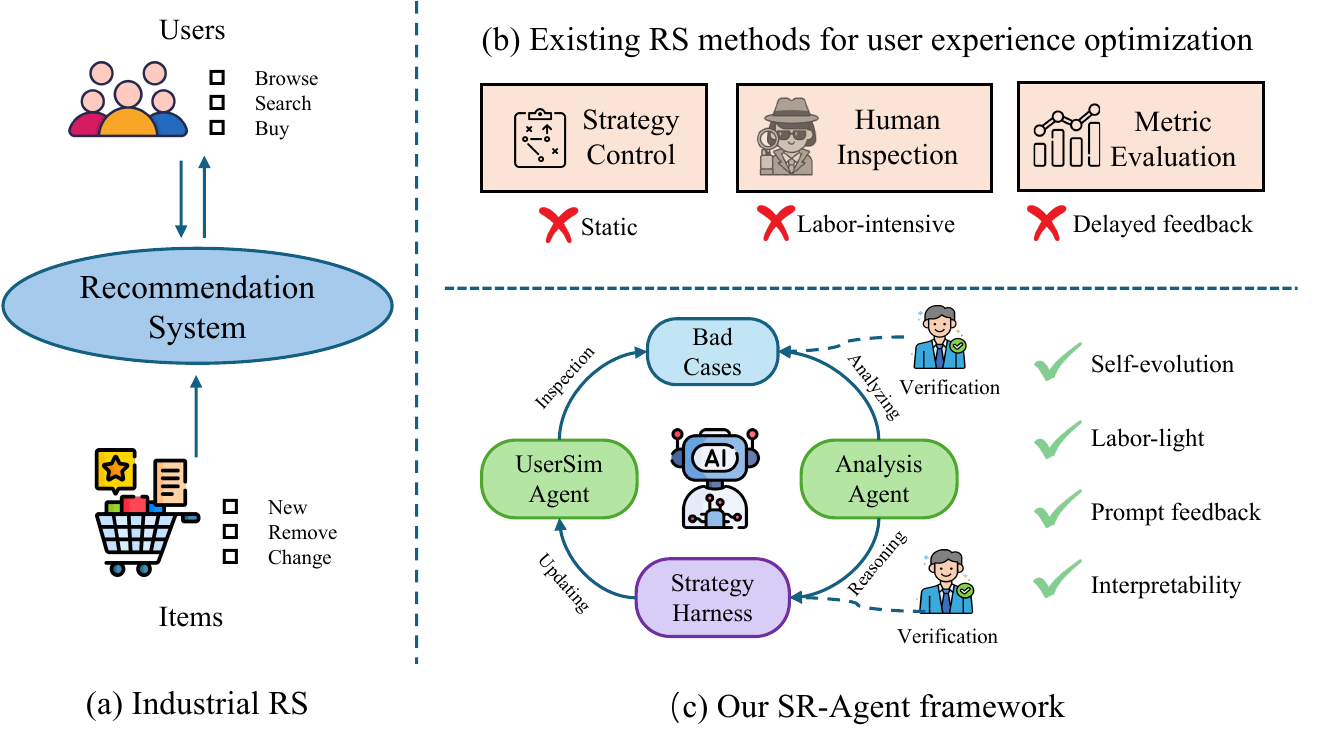}
  \caption{Overview of the motivation and proposed SR-Agent framework. Existing maintenance of post-ranking strategies relies on static configurations, manual inspection, and delayed metric feedback. Our agentic framework closes the loop from bad-case discovery and diagnosis to bounded strategy updates, offline verification, and controlled online validation.}
  \label{fig:overview}
\end{figure}

% \textbf{Challenge 1: both item and user states are dynamic, breaking the static assumptions of existing strategies, leading to bad cases.} Existing post-ranking strategies rely on static item attributes, predefined similarity functions, and manually configured thresholds. This assumption fails on both sides. On the \emph{item side}, millions of products are continuously listed, delisted, repriced, and revised in images or titles, causing taxonomy assignments, similarity thresholds, and substitutability rules to become stale. On the \emph{user side}, tolerance for similarity, repetition, and category concentration shifts with intent, session state, and recent exposure. Prior work on temporal diversity and unexpectedness confirms that list value depends on how items are arranged over time and how they match users' evolving expectations~\cite{lathia2010temporal,adamopoulos2015unexpectedness}. Static strategies cannot adapt to evolving user needs and dynamic item attributes, limiting their effectiveness in improving user experience.

\textbf{Challenge 1: Static post-ranking strategies degrade as the recommendation environment evolves, leading to recurring experience bad cases.}
Industrial recommenders commonly employ post-ranking strategies to control category, shop, brand, price-range, item-similarity, and exposure patterns. These strategies are typically configured using predefined item attributes, item relations, and fixed thresholds derived from historical observations~\cite{chen2019top}. However, the environment in which they operate evolves continuously. As shown in Fig.~\ref{fig:overview}, on the item side, products are frequently introduced, removed, repriced, relabeled, or updated in their titles and visual content, causing previously established attributes, relations, and thresholds to become stale. On the user side, preferences for product comparison, diversity, and repeated exposure vary with evolving interests, shopping intent, session state, and recent interactions~\cite{wang2021survey}. As static configurations gradually become misaligned with the current environment, they may incorrectly suppress useful alternatives or fail to remove redundant and functionally substitutable items. Consequently, their effectiveness degrades over time, giving rise to recurring experience bad cases that require continual strategy refinement.

\textbf{Challenge 2: Human-intensive strategy refinement is slow, costly, and difficult to scale.}
When experience bad cases emerge, industrial systems typically rely on a multi-stage manual workflow to refine the corresponding post-ranking strategies. Product or operation specialists first inspect recommendation results and identify potential experience defects; domain experts then analyze their causes and communicate the findings to algorithm engineers; the engineers translate these case-level observations into adjustments to category mappings, thresholds, penalties, or other strategy configurations; and the updated results must be inspected and verified again before deployment. Although this workflow is well established, it requires repeated coordination across multiple roles, incurs substantial human and time costs, and leads to long refinement cycles. Moreover, the quality of bad-case diagnosis and strategy adjustment depends heavily on the experience of individual experts. The resulting knowledge is often scattered across isolated cases and difficult to preserve, reuse, and improve in subsequent iterations.

Although recent developments in large language models (LLMs) and LLM-based agents offer promising directions for automating parts of this workflow, none of them close the full loop of automated, self-evolving strategy refinement. 
A number of LLM-as-a-judge methods enable scalable, reference-free evaluation of complex outputs~\cite{liu2023geval,zheng2023mtbench}, while LLM-based user simulators model user preferences, feedback, and sequential behavior for offline evaluation and training~\cite{zhang2023agent4rec,yoon2024usersim,bougie2025simuser,wanyan2025temporal}. However, these approaches primarily produce evaluation signals and do not directly convert detected experience defects into executable strategy updates~\cite{zheng2023mtbench,shi2024positionbias,yoon2024usersim}.
More recent self-evolving systems further use agents to propose, implement, and validate changes to recommendation models, objectives, or production code~\cite{kim2026selfevolverec,wang2026selfevolving,kuaishou2026agentx,mao2026stepfly}. Nevertheless, these systems primarily target recommendation decisions or broad model and code optimization. A scalable workflow that continuously connects user-perceived bad-case detection, cause diagnosis, reusable knowledge accumulation, bounded post-ranking strategy refinement, and deployment validation remains underexplored.

To address above issues, we introduce \textbf{SR-Agent}, an agentic framework that automates bad-case detection and diagnosis, accumulates reusable refinement knowledge, and supports bounded and verifiable updates to \emph{post-ranking strategies}.
Concretely, the \textbf{UserSim agent} inspects recommendation pages and session sequences through staged skills for exposure-pattern analysis, item-relation inference, and context-conditioned experience assessment, and outputs structured bad cases. The \textbf{Analysis agent}  explains each bad case by identifying the underlying relation type and failure cause, dynamic item substitution, taxonomy mismatch, loose thresholds, dense session exposure, etc, and abstracts recurring failure patterns into a reusable Diagnosis Memory, while the underlying bad cases are retained in Bad-Case Memory for retesting. The \textbf{Strategy refinement harness} maps these diagnoses to a predefined strategy action space and generates candidate updates through four typed actions: relation-threshold adjustment, redundancy-suppression adjustment, exposure-limit adjustment, and authorized taxonomy correction. Candidate updates must pass bad-case retesting, traffic replay, human verification, and online A/B validation before deployment.

Our main contributions are summarized as follows:

\begin{itemize}
  \item We formulate post-ranking strategy maintenance as an automated and self-evolving closed loop from experience inspection and diagnosis to bounded refinement and verification. By running continuously, the loop enables deployed strategies to adapt to the evolving online environment. 

  \item We develop {SR-Agent}, which combines two reasoning agents with a guarded refinement harness to automate bad-case analysis, accumulate reusable knowledge, and safely update post-ranking strategies, replacing the traditionally human-intensive workflow.

  \item Continuous experiments on Kuaishou, a large-scale e-commerce  platform, demonstrate that SR-Agent improves order volume by 0.71\%, browsing depth by 0.34\%, and clicked-category diversity by 0.48\%, while shortening the strategy refinement cycle and lowering costs.
\end{itemize}

\section{Related Work}

\subsection{Post-Ranking Strategies for User Experience}

User experience in recommender systems extends beyond predictive accuracy. Accuracy-oriented evaluation can overlook perceived usefulness, satisfaction, choice difficulty, and trust~\cite{mcnee2006accurate,pu2011resque,pu2012userperspective}, while diversity, novelty, serendipity, and unexpectedness provide complementary perspectives on list- and session-level quality~\cite{kunaver2017diversity,duricic2023beyondaccuracy,adamopoulos2015unexpectedness}. To operationalize these objectives after ranking, industrial recommenders employ \emph{post-ranking strategies}: rule-based or parameterized controls that modify the final list without changing the base ranker. Such strategies include category or shop quotas, similarity suppression, exposure caps, and algorithmic diversification based on MMR~\cite{carbonell1998mmr}, topic diversification~\cite{ziegler2005improving}, DPP~\cite{liu2022dpprec,wilhelm2018youtubedpp}, or temporal diversity~\cite{lathia2010temporal}.
Existing post-ranking strategies typically operate on predefined taxonomy labels, item representations, similarity functions, and manually configured thresholds. These signals are useful proxies for item relations, but they do not by themselves determine whether a displayed relation constitutes useful comparison, redundancy, or fatigue under a particular user and session context. Prior work primarily studies how to design or optimize a post-ranking method for producing an improved list; considerably less attention has been paid to maintaining deployed strategies after user-perceived failures emerge in live traffic. SR-Agent addresses this operational problem by detecting bad cases, diagnosing why the current strategies missed them, and translating structured diagnoses into bounded, auditable updates, therefore refining existing post-ranking strategies.

\subsection{LLM-Based User Simulation in Recommender Systems}

LLM-based user simulators provide a scalable means of modeling user feedback and interactions for the evaluation and training of recommender systems. Early work such as Agent4Rec constructs generative user agents with profile, memory, and action modules to simulate page-level recommendation interactions~\cite{zhang2023agent4rec}. For conversational recommendation, Yoon et al.~\cite{yoon2024usersim} introduce an evaluation protocol covering item selection, binary and open-ended preference expression, recommendation requests, and feedback, and show that LLM simulators can still deviate systematically from human behavior. Zhang et al.~\cite{zhang2025llm} instead model explicit user engagement with recommended items by combining LLM-derived preference logic with a statistical model, producing simulated interactions for reinforcement-learning-based recommender training. RecUserSim improves the realism and diversity of conversational user simulation through profile, memory, bounded-rationality-inspired action, and response-refinement modules, while additionally producing explicit ratings for quantitative evaluation~\cite{10.1145/3701716.3715258}. Moving beyond static user profiles, DyTA4Rec incorporates dynamic profile updating, temporally enhanced prompting, and adaptive aggregation to capture evolving user behavior~\cite{wanyan2025temporal}. SimUSER further integrates persona, memory, perception, and behavior modules to support recommender evaluation and simulator-guided refinement~\cite{bougie2025simuser}.
These studies primarily simulate user responses, engagement, or interaction trajectories for recommender evaluation and training. UserSim in SR-Agent instead combines user simulation with an LLM-as-a-judge mechanism: conditioned on the user's behavior history and the current recommendation context, the LLM adopts the user's perspective to assess realized lists and sessions for potential experience defects. Unlike a generic LLM judge that evaluates an output against a task-level rubric~\cite{liu2023geval,zheng2023mtbench}, UserSim performs context-aware experience assessment and produces evidence-grounded bad cases rather than a single aggregate score. These assessments are quality-controlled through human audits and provide actionable inputs for subsequent strategy diagnosis and refinement.

\subsection{Agentic Recommender Systems}

LLM-based agents have extended recommender systems from one-shot prediction toward interactive and reasoning-intensive recommendation. Existing studies employ agent capabilities such as memory, planning, tool use, and multi-agent coordination for preference elicitation, conversational recommendation, user--item reasoning, and explanation~\cite{huang2023interecagent,zhang2023agentcf,gao2024recmind,ma2025agentrec,park2025madrec}. For instance, Self-EvolveRec integrates user simulation with model diagnosis to provide directional feedback for iterative recommendation-model evolution~\cite{kim2026selfevolverec}. Industrial systems further employ LLM agents as machine-learning engineers to optimize model architectures, learning algorithms, and reward functions through offline and online feedback loops~\cite{wang2026selfevolving}, while AgentX automates the broader research cycle from hypothesis generation and code modification to online experimentation and knowledge accumulation~\cite{kuaishou2026agentx}. These systems demonstrate the potential of agents to close the loop of recommender optimization, but mainly focus on evolving recommendation models or automating the model-development workflow~\cite{lin2026autonomous}. In contrast, SR-Agent deliberately narrows self-evolution to a typed, bounded strategy layer, trading expressive power for auditability, fast validation, and low-risk deployment in always-on production maintenance.

\begin{figure}[t]
  \centering
  \includegraphics[width=\linewidth]{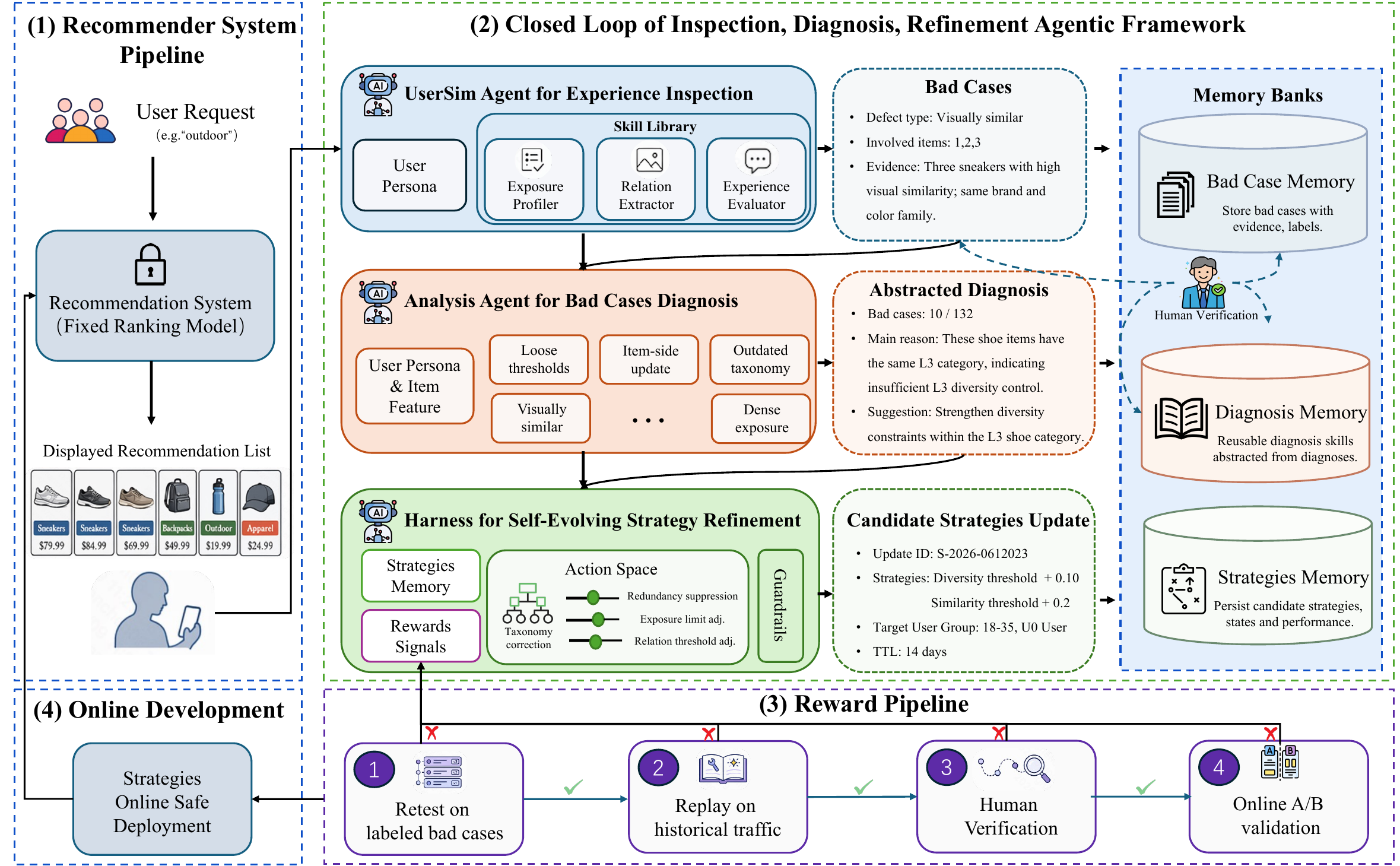}
  \caption{Overview of the proposed agentic framework. The base recommender produces displayed recommendation lists with a fixed ranking model. The agentic loop then performs user-perspective inspection, bad-case diagnosis, bounded strategy refinement, memory update, reward validation, and safe online deployment.}
  \label{fig:method-overview}
\end{figure}

\section{Method}

\subsection{Preliminaries and Problem Setup}
\label{sec:prelim}

We consider a production recommender in which a base ranker produces, for a user $u$ under serving context $c$, a scored candidate pool that is processed by a set of post-ranking strategies, jointly represented by $\pi_S$, before display. $\pi_S$ is parameterized by a strategy state $S\in\mathcal{S}$ that encodes diversity thresholds, similarity suppressions, item/relation penalties, exposure caps, and item groupings. The displayed list $L=\pi_S(u,c)=(i_1,\ldots,i_K)$ is what the user actually observes.

SR-Agent leaves the base ranker frozen and refines only $\pi_S$. Given traffic $\mathcal{T}=\{(u,L,c)\}$, deployed state $S$, logged history $\mathcal{D}_{\text{hist}}$, and a fixed typed action space $\mathcal{A}$, the loop produces a sequence of candidate updates $\{a_t\}\subset\mathcal{A}$ that must pass a guarded admission pipeline before being applied to $S$.  SR-Agent retrieves accumulated experience from three memories to inform subsequent inspection, diagnosis, and strategy refinement.

\subsection{Framework Overview}

Figure~\ref{fig:method-overview} illustrates the data flow of SR-Agent. The left part shows the production recommender pipeline: a user request is processed by the existing recommender system, whose ranking model remains fixed, to produce a displayed recommendation list. SR-Agent operates on this list without replacing the base recommender. The middle part presents the closed agentic loop. Conditioned on the user persona, the UserSim agent inspects the displayed list using rule-based, visual, and intent-reasoning skills and produces structured bad cases. The Analysis agent abstracts these cases into diagnoses that describe the underlying failure causes and corresponding refinement directions. Based on the diagnoses, current strategy states, and historical reward signals, the strategy-refinement harness generates bounded candidate strategy updates. The right part maintains the resulting bad cases, diagnoses, and strategy records in three memory banks. Finally, the bottom reward pipeline evaluates each candidate through labeled bad-case retesting, historical-traffic replay, human verification, and online A/B testing before safe deployment. Validation outcomes are fed back into the framework to guide subsequent refinements.

\subsection{UserSim Agent}

Aggregate performance metrics may fail to capture list- and
session-level experience defects: individually relevant items can
still form a repetitive, homogeneous, or fatiguing display. UserSim
therefore approximates how a user perceives the displayed results,
rather than simulating a complete trajectory of future user actions.
Given a user $u$, a displayed list $L$, and serving context $c$, it
constructs an inspection context from the user's recent behaviors and
exposure history, together with item attributes, category hierarchy,
product images and titles, shop, and price information.

UserSim examines this context through three complementary inspection
skills that are invoked in a staged manner:

\begin{itemize}
  \item The \textbf{Exposure Profiler skill} identifies observable list- and
  session-level patterns, including repeated exposure and excessive
  concentration over categories, shops, and price bands. It
  characterizes what the user repeatedly encounters.

  \item The \textbf{Relation Extractor skill } identifies
  functional relations among the displayed items, including relations
  not captured by existing taxonomy labels. It provides evidence of
  why a group of items may be perceived as similar, substitutable, or
  complementary.

  \item The \textbf{Experience Evaluator skill} integrates the preceding
  evidence with the user's current intent, recent behavior, and
  exposure history. It determines whether the observed recommendation list
  support useful comparison or instead constitute redundancy,
  homogeneity, or browsing fatigue.
\end{itemize}

Formally, the \emph{Exposure Profiler} and \emph{Relation Extractor}
produce exposure evidence $e_b^{\mathrm{exp}}$ and relation evidence
$e_b^{\mathrm{rel}}$, respectively. The \emph{Experience Evaluator}
then produces the defect type $y$, the involved items
$I_b\subseteq L$, context-conditioned evidence $e_b^{\mathrm{ctx}}$,
and confidence $q_b$:
\begin{equation}
  (y,\,I_b,\,e_b^{\mathrm{ctx}},\,q_b)
  =
  \operatorname{ExpEval}\!\bigl(u,\,c,\,e_b^{\mathrm{exp}},\,e_b^{\mathrm{rel}}\bigr),
\end{equation}
where $\operatorname{ExpEval}$ denotes the Experience Evaluator. It
integrates the exposure and relation evidence with the user's current
intent $u$, recent behavior, and exposure history $c$ to yield the
final experience judgment. The resulting bad case is stored as
\begin{equation}
  b=\bigl(y,\,I_b,\,e_b,\,q_b\bigr),\qquad
  e_b=\{e_b^{\mathrm{exp}},\,e_b^{\mathrm{rel}},\,e_b^{\mathrm{ee}}\}.
\end{equation}

\subsection{Analysis Agent}

UserSim decides \emph{whether} a displayed result is an experience bad
case; the Analysis agent decides \emph{why the deployed post-ranking
strategies let it through, and where to intervene}. The separation
matters because a single visible symptom---say, three visually similar
shoes in the top-6---can arise from very different operational causes:
missing relation signals, stale item groupings, miscalibrated
thresholds, or loose exposure control. Conflating them leads to the
wrong fix.

Given a bad-case cluster $B\subseteq\mathcal{M}_b$ (grouped by defect
type, involved items or categories, evidence patterns, and serving
scenario) and the deployed strategy state $S$, the Analysis agent runs
a structured chain-of-thought that narrows the diagnosis in three
stages, each conditioned on the previous:

\begin{itemize}
  \item \textbf{What do the bad cases have in common?} Consolidate the per-case evidence produced by UserSim across the cluster $B$ and  extract the shared patterns, the recurring item-relation type $z$ (visual or semantic similarity, functional substitution,
  complementarity, or repeated exposure), together with the co-occurring item, category, and context signals. This step turns a bag of noisy  per-case observations into a single, cluster-level hypothesis about the underlying item-level phenomenon, and filters out one-off noise  before any causal reasoning begins.

  \item \textbf{Why do the current strategies miss it?} Contrast the
  shared pattern under $z$ against the deployed strategy state $S$ to
  attribute an operational cause $h$: missing signal coverage, stale
  grouping, threshold miscalibration, insufficient exposure control, or
  a mismatch between strategy tolerance and user context.

  \item \textbf{How should we fix it?} Given
  $(z,h)$, infer the smallest defensible scope $s$, an item group,
  category subtree, serving scenario, or user segment, and a
  high-level refinement direction $v$ (tighten or relax a constraint,
  adjust a threshold or penalty, revise an authorized grouping).
\end{itemize}

Each diagnosis is recorded as
\begin{equation}
  d=(z,h,s,v,e_d),
\end{equation}
where $(z,h,s,v)$ carry the outputs of the three CoT stages and $e_d$
bundles the supporting bad cases, consolidated evidence, and the
relevant slice of $S$ for downstream retracing. In the continuously running
production loop, human reviewers randomly audit approximately 1\% of the bad
cases produced by UserSim and 1\% of the diagnoses produced by the Analysis
agent. This sampled quality-control audit is distinct from the guarded
validation required for candidate strategy updates before deployment.

\subsection{Strategy Refinement Harness}

The strategy-refinement harness provides a constrained interface
between diagnosis and online deployment. Its action space, output
schema, scope permissions, and deployment guardrails are fixed and
cannot be modified by the LLM. Its adaptive state is Strategy Memory
$\mathcal{M}_s$, which records previous proposals and their validation
outcomes.

\paragraph{Typed action space.}
Before the optimization loop, strategy owners configure and version
the typed action space $\mathcal{A}$ summarized in
Table~\ref{tab:action-space}. The LLM cannot emit serving code, alter
model objectives, target unauthorized scopes, or invoke operations
outside $\mathcal{A}$.

The action types address different user-experience failures.
A \emph{relation-threshold} action corrects overly permissive or
restrictive identification of similar or substitutable items.
A \emph{redundancy-suppression} action adjusts how strongly identified
relations affect the displayed list. An \emph{exposure-limit} action
controls excessive concentration over dimensions such as category,
shop, and price band. Finally, a \emph{taxonomy correction} repairs
authorized taxonomy mappings that prevent functionally equivalent
items from being treated consistently.

Figure~\ref{fig:method-overview} uses serving-side shorthand for these
schema-level actions: threshold updates correspond to relation-threshold
adjustments; item-side penalties and dispersion-strength changes correspond to
redundancy-suppression adjustments; segment-tolerance changes are implemented
through exposure-limit configurations over authorized scopes; and taxonomy
changes correspond to taxonomy corrections. The candidate card in the figure
shows resulting configured strategy values, whereas Table~\ref{tab:action-space}
specifies the admissible shift applied by any single candidate update.

\begin{table}[h]
  \centering
  \caption{Typed action space $\mathcal{A}$. Parameter bounds, target
    whitelists.}
  \label{tab:action-space}
  \small
  \renewcommand{\arraystretch}{1.15}
  \begin{tabular}{
    >{\centering\arraybackslash}p{0.23\linewidth}
    >{\raggedright\arraybackslash}p{0.3\linewidth}
    >{\raggedright\arraybackslash}p{0.45\linewidth}
  }
    \toprule
    Action type & Parameters & Admissible domain \\
    \midrule

    Relation threshold adjustment
      & Relation $r$; threshold shift $\Delta\tau_r$
      & $r\in\{\textsc{visual},\textsc{semantic},\textsc{substitute}\}$;\newline
        $\Delta\tau_r\in[-0.15,+0.15]$, step $0.01$ \\

    Redundancy suppression adjustment
      & Group $g$; strength shift $\Delta\lambda$
      & $g\in\mathcal{G}_{\mathrm{auth}}$;\newline
        $\Delta\lambda\in[-0.50,+0.50]$, step $0.05$ \\

    Exposure limit adjustment
      & Dimension $d$; horizon $h$; limit $k$
      & $d\in\{\textsc{category},\textsc{shop},\textsc{price-band}\}$;\newline
        $(h,k)\in\mathcal{C}_{\mathrm{cfg}}$ \\

    Taxonomy correction
      & Operation $o$; source node(s) $n$; target group $g$
      & $o\in\{\textsc{merge},\textsc{split},\textsc{reassign}\}$;\newline
        $(n,g)\in\mathcal{C}_{\mathrm{tax}}^{\mathrm{auth}}$ \\

    \bottomrule
  \end{tabular}
\end{table}

\paragraph{Candidate generation and validation.}
For each diagnosis $d$, the harness retrieves relevant precedents
\[
H=\operatorname{Retrieve}(d,\mathcal{M}_s)
\]
according to diagnosis, scope, and historical outcome. It then
generates a small candidate set
\begin{equation}
  C=G(d,S,H;\mathcal{A})\subseteq\mathcal{A},
  \qquad
  a=(\mathrm{id},s,t,p,E,T),
\end{equation}
where $s$ is the authorized scope, $t$ the action type, $p$ its bounded
parameters, $E$ the supporting evidence, and $T$ the time-to-live.

A deterministic validator rejects candidates with invalid parameters,
unauthorized scopes, unsupported operations, or incompatible strategy
versions. Compatible parameter updates may be composed and clipped to
their authorized bounds; conflicting updates and taxonomy operations
are escalated for strategy-owner review.

\paragraph{Guarded evaluation and memory update.}
Each valid candidate passes four stages. First, the harness invokes the
pre-registered strategy script associated with the candidate action on
diagnosis-matched cases from $\mathcal{M}_b$. The LLM supplies only
bounded parameters and cannot generate or modify the script. This
bad-case retest measures whether the targeted defect is corrected.
Second, the same script is replayed on the larger historical sample
$\mathcal{D}_{\mathrm{hist}}$ to estimate population-level coverage,
list changes, and unintended regressions. Thus, bad-case retesting
measures targeted effectiveness, whereas historical traffic replay
measures generality and safety beyond the selected cases. Candidates
that pass both offline stages proceed to strategy-owner review and
controlled online A/B testing. Their outcomes are recorded as
\begin{equation}
  r_t=
  \bigl(
    r_t^{\mathrm{case}},
    r_t^{\mathrm{replay}},
    r_t^{\mathrm{human}},
    r_t^{\mathrm{online}}
  \bigr).
\end{equation}
Failure at any stage stops the candidate, while an online guardrail
violation triggers rollback. The complete outcome is stored in
Strategy Memory:
\begin{equation}
  \mathcal{M}_s^{t+1}
  =
  \mathcal{M}_s^t
  \cup
  \{(d_t,a_t,r_t,o_t)\},
\end{equation}
where
$o_t\in\{\text{invalid},\text{rejected},\text{deployed},
\text{rolled-back}\}$.

Algorithm~\ref{alg:SR} summarizes the refinement loop.

\begin{algorithm}[t]
\caption{SR-Agent strategy-refinement loop}
\label{alg:SR}
\begin{algorithmic}[1]
\Require traffic $\mathcal{T}$, state $S$, memories
$\mathcal{M}_b,\mathcal{M}_d,\mathcal{M}_s$, history
$\mathcal{D}_{\mathrm{hist}}$, action space $\mathcal{A}$

\State $\mathcal{B}_{\text{new}}\gets\varnothing$
\For{$(u,L,c)\in\mathcal{T}$}
  \State $b\gets\Call{UserSim}{u,L,c}$
  \If{$b$ is valid and sufficiently confident}
    \State append $b$ to $\mathcal{M}_b$ and
    $\mathcal{B}_{\text{new}}$
  \EndIf
\EndFor

\For{cluster $B\in\Call{ClusterBadCases}{\mathcal{B}_{\mathrm{new}}}$}
  \State $d\gets\Call{Analysis}{B,\mathcal{M}_d}$
  \If{$d$ satisfies the diagnosis contract}
    \State append $d$ to $\mathcal{M}_d$
    \State $H\gets\Call{RetrievePrecedents}{d,\mathcal{M}_s}$
    \State $C\gets\Call{Harness}{d,S,H,\mathcal{A}}$

    \For{$a\in C$}
      \State $a'\gets\Call{ValidateAndResolve}{a,S,\mathcal{A}}$
      \If{$a'=\varnothing$}
        \State record $(d,a,\varnothing,\mathrm{invalid})$ in
        $\mathcal{M}_s$
      \Else
        \State $(r,o)\gets
        \Call{GuardedEvaluate}{a',\mathcal{M}_b,
        \mathcal{D}_{\mathrm{hist}}}$
        \If{$o=\mathrm{deployed}$}
          \State $S\gets\Call{Deploy}{S,a'}$
        \EndIf
        \State record $(d,a',r,o)$ in $\mathcal{M}_s$
      \EndIf
    \EndFor
  \EndIf
\EndFor

\State \Return updated $S,\mathcal{M}_b,\mathcal{M}_d,\mathcal{M}_s$
\end{algorithmic}
\end{algorithm}

\section{Experiments}

To evaluate SR-Agent in a real-world industrial setting, we deployed it on
Kuaishou's e-commerce platform, a large-scale, content-driven platform serving
tens of millions of daily active users. 
To isolate the effect of SR-Agent, the control group retained the existing
manually configured post-ranking strategies, whereas the treatment group kept
the base ranking model unchanged and allowed SR-Agent to refine only predefined,
bounded strategies. Within this deployment, we conducted a one-month online
A/B test, with 2.5\% of the traffic assigned to the treatment group.  During the test period, the deployment covered approximately 60 million users and 70 million items.
Our evaluation addresses the following research questions:
\begin{itemize}
    \item \textbf{RQ1}: Does SR-Agent improve online user experience and
    business metrics?
    \item \textbf{RQ2}: To what extent are the bad cases identified by UserSim
    validated by human reviewers?
    \item \textbf{RQ3}: How much does SR-Agent reduce case inspection and
    end-to-end strategy iteration time compared with the legacy manual workflow?
    \item \textbf{RQ4}: How do user-experience metrics evolve over successive
    iterations of the SR-Agent optimization loop?
\end{itemize}

\subsection{RQ1: Online A/B Test}

Table~\ref{tab:online-ab} summarizes the one-month online A/B test in terms of
user experience and order-related performance. SR-Agent improves all reported
metrics. It increases the number of clicked categories, browsing depth, and
clicks, while also improving order volume and gross merchandise volume (GMV).
These results demonstrate that SR-Agent enhances the browsing experience while
delivering positive business outcomes.

\begin{table*}[h]
  \centering
  \caption{One-month online A/B results of SR-Agent. Values are relative changes compared with the control group.}
  \label{tab:online-ab}
  \small
  \begin{tabular}{llr}
    \toprule
    Metric group & Metric & Relative change \\
    \midrule
    User experience & Clicked categories per user (1-day) & $+0.425\%$ \\
    User experience & Clicked categories per user (7-day) & $+0.482\%$ \\
    User experience & Browsing depth & $+0.342\%$ \\
    User experience & Number of clicks & $+0.554\%$ \\
    Order-related & Order volume & $+0.707\%$ \\
    Order-related & Gross merchandise volume (GMV) & $+0.461\%$ \\
    \bottomrule
  \end{tabular}
\end{table*}

\subsection{RQ2: Effectiveness of Bad-Case Detection}

% NOTE: The 36 manually identified cases are treated as a workflow baseline,
% whereas 91\% and 95\% come from a separate validation of agent-flagged cases.
% If 36 is intended as exhaustive ground truth, the reported values conflict.
We evaluate bad-case detection on 500 recommendation cases sampled from
production traffic.  As shown in Table~\ref{tab:correctness},Manual inspection identifies 36 bad cases. UserSim without
user intent detects 114 cases with a 91\% validation rate, while UserSim with
user intent detects 142 cases with a 95\% validation rate.
Compared with manual inspection, although UserSim produces a small number of false positives, it identifies substantially more valid bad cases than manual inspection, demonstrating broader
coverage and greater effectiveness.

\begin{table}[h]
  \centering
  \caption{Bad-case detection results on 500 recommendation cases.}
  \label{tab:correctness}
  \small
  \begin{tabular}{lcc}
    \toprule
    Method & Detected cases & Validation rate \\
    \midrule
    Manual inspection & 36 & 100\% \\
    UserSim without user intent & 114 & 91\% \\
    UserSim with user intent & 142 & 95\% \\
    \bottomrule
  \end{tabular}
\end{table}

\subsection{RQ3: Human-in-the-Loop Operational Efficiency}

We compare SR-Agent with the legacy workflow in terms of model usage,
stage-level time, and end-to-end iteration time. All LLM-based
components use Claude Sonnet 4.6, and one complete loop consumes 36 Million
aggregate tokens across all agent calls. Importantly, SR-Agent does not remove
human participation: it automates large-scale inspection, diagnosis, and
candidate generation, while human reviewers randomly audit approximately 1\%
of generated bad cases and diagnoses and review candidate strategy updates
before online deployment.

\begin{table*}[h]
  \centering
  \caption{Operational efficiency of one strategy-refinement loop. Stage-level results report automated runtime and human effort separately; end-to-end iteration denotes elapsed wall-clock time.}
  \label{tab:operational-efficiency}
  \small
  \begin{tabular}{@{}p{0.30\textwidth}p{0.22\textwidth}p{0.36\textwidth}@{}}
    \toprule
    Dimension & Legacy workflow & SR-Agent (automation + human) \\
    \midrule
    LLM and capacity & No LLM & Claude Sonnet 4.6 \\
    Average Tokens per loop & 0 & 36 Million \\
    Inspection of 500 cases & 7 days & 50 min automation + 1 reviewer-day \\
    Analysis of 500 bad cases & 2 days & 90 min automation + 1 reviewer-day \\
    Strategy update & 2 weeks & 3 h automation + 1 reviewer-day \\
    End-to-end iteration & 2--3 weeks & 3--5 elapsed days \\
    \bottomrule
  \end{tabular}
\end{table*}

\subsection{RQ4: Bad-Case Resolution and Online Trend}

We track how SR-Agent improves over ten successive strategy updates, each
produced by one invocation of the refinement harness on a single bounded
strategy. For update $t$, we report the \emph{bad-case resolution rate}: the
percentage of targeted bad cases whose defect disappears in post-update
retesting while passing the replay guardrails. In parallel, we monitor the
one-month online rollout (June 15--July 15, 2026), measuring the relative
change in number of clicks and browsing depth against the baseline.

\begin{figure}[h]
  \centering
  \includegraphics[width=0.92\linewidth]{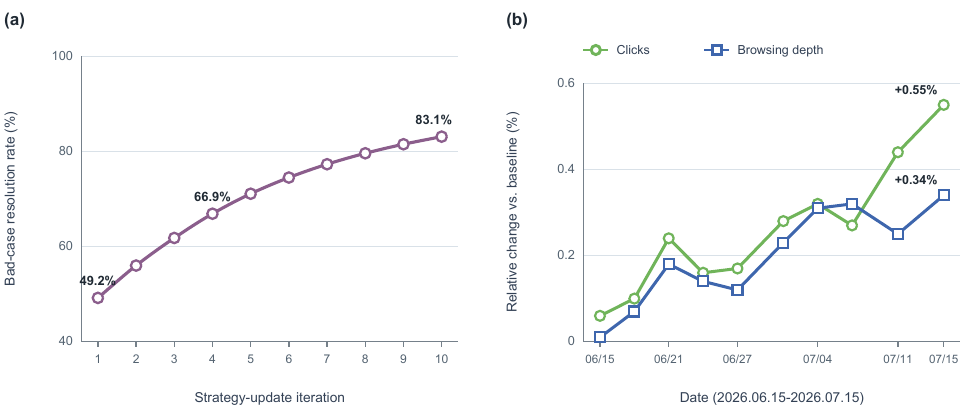}
  \caption{Results over ten strategy updates. (a) Per-update bad-case
  resolution rate of the refinement harness. (b) Relative change in number of
  clicks and browsing depth during the one-month online rollout, versus the
  pre-rollout baseline.}
  \label{fig:rq4-combined}
\end{figure}

As shown in Figure~\ref{fig:rq4-combined}(a), the resolution rate climbs from
49.2\% at the first update to 83.1\% at the tenth (+33.9 pp), with diminishing
per-update gains as common failures are fixed early and later updates target
rarer, harder cases. Figure~\ref{fig:rq4-combined}(b) shows a matching online
trend: despite short-term fluctuation from traffic composition and user
behavior, both metrics rise overall, reaching +0.55\% clicks and +0.34\%
browsing depth by the end of the rollout. Together, these results indicate that
the harness's growing ability to resolve bad cases translates into a
progressively better online browsing experience.

\subsection{Case Study}

\begin{figure}[t]
  \centering
  \includegraphics[width=0.9\linewidth]{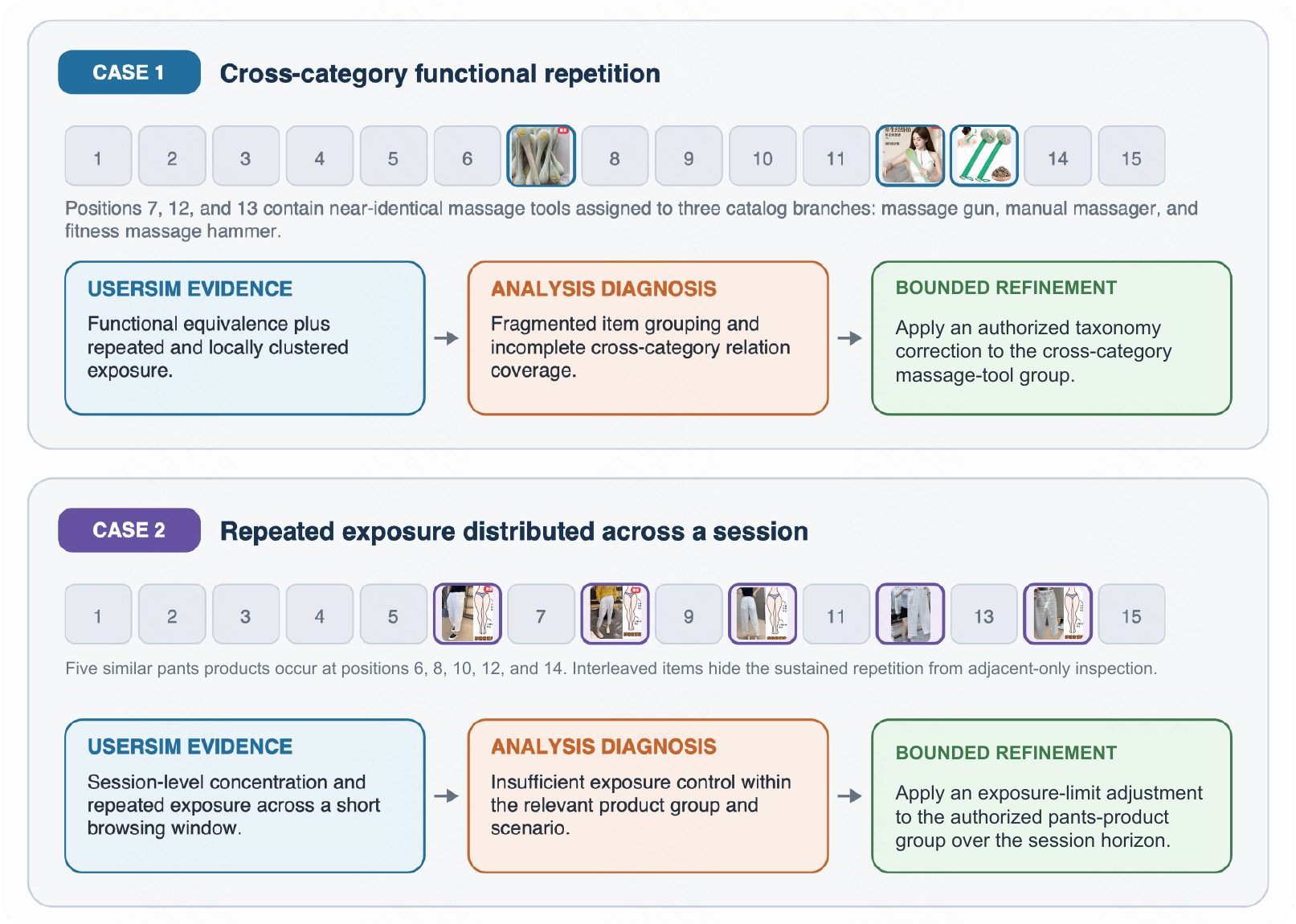}
  \caption{Two cases processed by SR-Agent. The first requires cross-category item-relation inference, whereas the second requires session-level exposure analysis. In both cases, UserSim evidence is converted into an actionable diagnosis and mapped to a bounded post-ranking strategy refinement.}
  \label{fig:case-study}
\end{figure}

We further examine two representative production cases, shown in
Figure~\ref{fig:case-study}, to illustrate how SR-Agent connects
user-perceived defects to actionable strategy diagnoses.

\textbf{Case 1: Cross-category repetition of functionally equivalent items.}
In the recommendation list, three nearly identical handheld massage products
appear at positions 7, 12, and 13. Although they address the same user need,
they are assigned to different catalog branches---massage hammers, manual
massagers, and fitness massage tools. Consequently, the deployed category-based
post-ranking strategy fails to recognize them as redundant exposures. By
jointly considering item semantics, functional relations, and list context,
UserSim identifies both repeated and locally clustered exposure. The Analysis
agent attributes the defect to incomplete relation coverage caused by
fragmented item groupings and localizes its scope to the corresponding
cross-category massage-tool group. Based on this diagnosis, the harness can
generate an authorized taxonomy-correction candidate that merges or reassigns
the involved catalog nodes to the cross-category massage-tool group.

\textbf{Case 2: Repeated exposure distributed across a session.}
In another item session, five highly similar pants products appear at
positions 6, 8, 10, 12, and 14. Because the items are interleaved with unrelated
products, no single adjacent pair fully captures the user-facing repetition,
and the online strategies with window 2 are not activated. By aggregating exposure evidence
over the session, UserSim identifies a sustained concentration pattern that
would be missed by purely local inspection. The Analysis agent attributes the
defect to insufficient exposure control within the relevant product group and
recommends tightening the exposure limit for the authorized pants-product
group over a configured session horizon. The harness expresses this direction
as an exposure-limit adjustment and submits the resulting candidate to
the guarded validation pipeline.

% \section{Limitations and Future Work}

% SR-Agent is intentionally conservative in scope: the typed action space $\mathcal{A}$ is bounded and the underlying LLM is not fine-tuned online, which secures safety and rollback but also means that defects requiring changes beyond post-ranking, such as new signals or ranking-model update, fall outside the current loop. Our empirical validation covers two large-scale e-commerce feeds; broader generalization to other post-ranking layouts and domains, together with principled extensions of $\mathcal{A}$ and tighter reward-shaping over $\mathcal{M}_s$, are natural next steps.

\section{Conclusion}

We presented SR-Agent, an agentic framework that closes the loop from user-perceived bad-case detection to bounded, reward-validated refinement of post-ranking strategies in industrial e-commerce recommendation. A UserSim agent surfaces evidence-grounded bad cases, an Analysis agent consolidates them into reusable diagnoses, and a constrained refinement harness turns each diagnosis into typed, bounded actions that are admitted only after a four-stage reward pipeline with reversible rollback. This decomposition keeps every online change interpretable, auditable, and safe to revert, which is what makes continuous self-refinement acceptable in production.
Deployed on Kuaishou, SR-Agent delivers on both axes we set out to improve. On effectiveness, a one-month online A/B test yields consistent gains in clicked-category diversity, browsing depth, clicks, and order volume. On efficiency, SR-Agent significantly compresses the end-to-end strategy-refinement cycle that previously required cross-role manual work, enabling faster iteration while retaining lightweight human audits.
In the future, we plan to generalize the agentic loop to other components of industrial RS, such as cold-start boosting and multi-channel retrieval quotas.

\section*{Declaration on Generative AI}

During the preparation of this work, the authors only used GenAI for language polishing, including grammar correction and improvements to clarity and fluency. The authors reviewed and edited all suggested revisions and take full responsibility for the content of this paper.

\bibliography{references}

@inproceedings{carbonell1998mmr,
  title={The use of MMR, diversity-based reranking for reordering documents and producing summaries},
  author={Carbonell, Jaime and Goldstein, Jade},
  booktitle={Proceedings of the 21st annual international ACM SIGIR conference on Research and development in information retrieval},
  pages={335--336},
  year={1998}
}

@inproceedings{mcnee2006accurate,
  title={Being accurate is not enough: how accuracy metrics have hurt recommender systems},
  author={McNee, Sean M and Riedl, John and Konstan, Joseph A},
  booktitle={CHI'06 extended abstracts on Human factors in computing systems},
  pages={1097--1101},
  year={2006}
}

@inproceedings{10.1145/3701716.3715258,
  title={Recusersim: A realistic and diverse user simulator for evaluating conversational recommender systems},
  author={Chen, Luyu and Dai, Quanyu and Zhang, Zeyu and Feng, Xueyang and Zhang, Mingyu and Tang, Pengcheng and Chen, Xu and Zhu, Yue and Dong, Zhenhua},
  booktitle={Companion Proceedings of the ACM on Web Conference 2025},
  pages={133--142},
  year={2025}
}

@article{lin2026autonomous,
  title         = {Autonomous Information Seeking: A Roadmap for Agentic Recommender Systems},
  author        = {Lin, Xinyu and Deldjoo, Yashar and Dai, Sunhao and Bao, Honghui and Ye, Xiaopeng and Nazary, Fatemeh and Wang, Wenjie and Di Noia, Tommaso and Xu, Jun and Chua, Tat-Seng},
  year          = {2026},
  journal       = {arXiv preprint arXiv:2607.04433},
  archivePrefix = {arXiv},
}

@inproceedings{zhang2025llm,
  title={Llm-powered user simulator for recommender system},
  author={Zhang, Zijian and Liu, Shuchang and Liu, Ziru and Zhong, Rui and Cai, Qingpeng and Zhao, Xiangyu and Zhang, Chunxu and Liu, Qidong and Jiang, Peng},
  booktitle={Proceedings of the AAAI Conference on Artificial Intelligence},
  volume={39},
  number={12},
  pages={13339--13347},
  year={2025}
}

@article{mao2026stepfly,
  title={StepFly: Agentic Troubleshooting Guide Automation for Incident Diagnosis},
  author={Mao, Jiayi and Li, Liqun and Gao, Yanjie and Peng, Zegang and He, Shilin and Zhang, Chaoyun and Qin, Si and Khalid, Samia and Lin, Qingwei and Rajmohan, Saravan and others},
  journal={Proceedings of the ACM on Software Engineering},
  volume={3},
  number={FSE},
  pages={3070--3092},
  year={2026},
  publisher={ACM New York, NY, USA}
}

@article{wang2021survey,
  title={A survey on session-based recommender systems},
  author={Wang, Shoujin and Cao, Longbing and Wang, Yan and Sheng, Quan Z and Orgun, Mehmet A and Lian, Defu},
  journal={ACM Computing Surveys (CSUR)},
  volume={54},
  number={7},
  pages={1--38},
  year={2021},
  publisher={ACM New York, NY, USA}
}

@inproceedings{chen2019top,
  title={Top-k off-policy correction for a REINFORCE recommender system},
  author={Chen, Minmin and Beutel, Alex and Covington, Paul and Jain, Sagar and Belletti, Francois and Chi, Ed H},
  booktitle={Proceedings of the twelfth ACM international conference on web search and data mining},
  pages={456--464},
  year={2019}
}

@inproceedings{song2025user,
  title={User-centric internal tools in e-commerce: Enhancing operational efficiency through ai integration},
  author={Song, Xue},
  booktitle={Proceedings of the 2025 International Conference on Economic Management and Big Data Application},
  pages={702--706},
  year={2025}
}

@inproceedings{adaji2017towards,
  title={Towards improving e-commerce users experience using personalization \& persuasive technology},
  author={Adaji, Ifeoma},
  booktitle={Proceedings of the 25th Conference on User Modeling, Adaptation and Personalization},
  pages={318--321},
  year={2017}
}

@inproceedings{ziegler2005improving,
  title={Improving recommendation lists through topic diversification},
  author={Ziegler, Cai-Nicolas and McNee, Sean M and Konstan, Joseph A and Lausen, Georg},
  booktitle={Proceedings of the 14th international conference on World Wide Web},
  pages={22--32},
  year={2005}
}

@article{kunaver2017diversity,
  title={Diversity in recommender systems--A survey},
  author={Kunaver, Matev{\v{z}} and Po{\v{z}}rl, Toma{\v{z}}},
  journal={Knowledge-based systems},
  volume={123},
  pages={154--162},
  year={2017},
  publisher={Elsevier}
}

@article{duricic2023beyondaccuracy,
  title={Beyond-accuracy: a review on diversity, serendipity, and fairness in recommender systems based on graph neural networks},
  author={Duricic, Tomislav and Kowald, Dominik and Lacic, Emanuel and Lex, Elisabeth},
  journal={Frontiers in big data},
  volume={6},
  pages={1251072},
  year={2023},
  publisher={Frontiers}
}

@inproceedings{liu2022dpprec,
  title={Determinantal point process likelihoods for sequential recommendation},
  author={Liu, Yuli and Walder, Christian and Xie, Lexing},
  booktitle={Proceedings of the 45th international ACM SIGIR conference on research and development in information retrieval},
  pages={1653--1663},
  year={2022}
}

@inproceedings{lathia2010temporal,
  title={Temporal diversity in recommender systems},
  author={Lathia, Neal and Hailes, Stephen and Capra, Licia and Amatriain, Xavier},
  booktitle={Proceedings of the 33rd international ACM SIGIR conference on Research and development in information retrieval},
  pages={210--217},
  year={2010}
}

@article{adamopoulos2015unexpectedness,
  title={On unexpectedness in recommender systems: Or how to better expect the unexpected},
  author={Adamopoulos, Panagiotis and Tuzhilin, Alexander},
  journal={ACM Transactions on Intelligent Systems and Technology (TIST)},
  volume={5},
  number={4},
  pages={1--32},
  year={2014},
  publisher={ACM New York, NY, USA}
}

@inproceedings{pu2011resque,
  title={A user-centric evaluation framework for recommender systems},
  author={Pu, Pearl and Chen, Li and Hu, Rong},
  booktitle={Proceedings of the fifth ACM conference on Recommender systems},
  pages={157--164},
  year={2011}
}

@article{pu2012userperspective,
  title={Evaluating recommender systems from the user’s perspective: survey of the state of the art},
  author={Pu, Pearl and Chen, Li and Hu, Rong},
  journal={User Modeling and User-Adapted Interaction},
  volume={22},
  number={4},
  pages={317--355},
  year={2012},
  publisher={Springer}
}

@inproceedings{liu2023geval,
  title={G-eval: NLG evaluation using gpt-4 with better human alignment},
  author={Liu, Yang and Iter, Dan and Xu, Yichong and Wang, Shuohang and Xu, Ruochen and Zhu, Chenguang},
  booktitle={Proceedings of the 2023 conference on empirical methods in natural language processing},
  pages={2511--2522},
  year={2023}
}

@article{zheng2023mtbench,
  title={Judging llm-as-a-judge with mt-bench and chatbot arena},
  author={Zheng, Lianmin and Chiang, Wei-Lin and Sheng, Ying and Zhuang, Siyuan and Wu, Zhanghao and Zhuang, Yonghao and Lin, Zi and Li, Zhuohan and Li, Dacheng and Xing, Eric and others},
  journal={Advances in neural information processing systems},
  volume={36},
  pages={46595--46623},
  year={2023}
}

@inproceedings{shi2024positionbias,
  title={Judging the judges: A systematic study of position bias in llm-as-a-judge},
  author={Shi, Lin and Ma, Chiyu and Liang, Wenhua and Diao, Xingjian and Ma, Weicheng and Vosoughi, Soroush},
  booktitle={Proceedings of the 14th International Joint Conference on Natural Language Processing and the 4th Conference of the Asia-Pacific Chapter of the Association for Computational Linguistics},
  pages={292--314},
  year={2025}
}

@inproceedings{zhang2023agent4rec,
  title={On generative agents in recommendation},
  author={Zhang, An and Chen, Yuxin and Sheng, Leheng and Wang, Xiang and Chua, Tat-Seng},
  booktitle={Proceedings of the 47th international ACM SIGIR conference on research and development in Information Retrieval},
  pages={1807--1817},
  year={2024}
}

@article{huang2023interecagent,
  title={Recommender ai agent: Integrating large language models for interactive recommendations},
  author={Huang, Xu and Lian, Jianxun and Lei, Yuxuan and Yao, Jing and Lian, Defu and Xie, Xing},
  journal={ACM Transactions on Information Systems},
  volume={43},
  number={4},
  pages={1--33},
  year={2025},
  publisher={ACM New York, NY}
}

@inproceedings{zhang2023agentcf,
  title={Agentcf: Collaborative learning with autonomous language agents for recommender systems},
  author={Zhang, Junjie and Hou, Yupeng and Xie, Ruobing and Sun, Wenqi and McAuley, Julian and Zhao, Wayne Xin and Lin, Leyu and Wen, Ji-Rong},
  booktitle={Proceedings of the ACM Web Conference 2024},
  pages={3679--3689},
  year={2024}
}

@inproceedings{yoon2024usersim,
  title={Evaluating large language models as generative user simulators for conversational recommendation},
  author={Yoon, Se-eun and He, Zhankui and Echterhoff, Jessica and McAuley, Julian},
  booktitle={Proceedings of the 2024 Conference of the North American Chapter of the Association for Computational Linguistics: Human Language Technologies (Volume 1: Long Papers)},
  pages={1490--1504},
  year={2024}
}

@inproceedings{bougie2025simuser,
  title={Simuser: Simulating user behavior with large language models for recommender system evaluation},
  author={Bougie, Nicolas and Watanabe, Narimawa},
  booktitle={Proceedings of the 63rd Annual Meeting of the Association for Computational Linguistics (Volume 6: Industry Track)},
  pages={43--60},
  year={2025}
}

@inproceedings{wanyan2025temporal,
  title={Temporal-Aware User Behaviour Simulation with Large Language Models for Recommender Systems},
  author={Wanyan, Xinye and Hettiachchi, Danula and Ma, Chenglong and Xu, Ziqi and Chan, Jeffrey},
  booktitle={Proceedings of the 34th ACM International Conference on Information and Knowledge Management},
  pages={5335--5339},
  year={2025}
}

@article{kim2026selfevolverec,
  title={Self-EvolveRec: Self-Evolving Recommender Systems with LLM-based Directional Feedback},
  author={Kim, Sein and Park, Sangwu and Kang, Hongseok and Kim, Wonjoong and Seo, Jimin and In, Yeonjun and Yoon, Kanghoon and Park, Chanyoung},
  journal={arXiv preprint arXiv:2602.12612},
  year={2026}
}

@article{wang2026selfevolving,
  title={Self-evolving recommendation system: End-to-end autonomous model optimization with LLM agents},
  author={Wang, Haochen and Wu, Yi and Chang, Daryl and Wei, Li and Heldt, Lukasz},
  journal={arXiv preprint arXiv:2602.10226},
  year={2026}
}

@article{ma2025agentrec,
  title={AgentRec: Next-Generation LLM-Powered Multi-Agent Collaborative Recommendation with Adaptive Intelligence},
  author={Ma, Bo and Li, Hang and Hu, ZeHua and Gui, XiaoFan and Liu, LuYao and Lau, Simon},
  journal={arXiv preprint arXiv:2510.01609},
  year={2025}
}

@article{park2025madrec,
  title={MADREC: A Multi-Aspect Driven LLM Agent for Explainable and Adaptive Recommendation},
  author={Park, Jiin and Kim, Misuk},
  journal={arXiv preprint arXiv:2510.13371},
  year={2025}
}

@inproceedings{gao2024recmind,
  title={Recmind: Large language model powered agent for recommendation},
  author={Wang, Yancheng and Jiang, Ziyan and Chen, Zheng and Yang, Fan and Zhou, Yingxue and Cho, Eunah and Fan, Xing and Lu, Yanbin and Huang, Xiaojiang and Yang, Yingzhen},
  booktitle={Findings of the Association for Computational Linguistics: NAACL 2024},
  pages={4351--4364},
  year={2024}
}

@article{chen2018fastdpp,
  title={Fast greedy map inference for determinantal point process to improve recommendation diversity},
  author={Chen, Laming and Zhang, Guoxin and Zhou, Eric},
  journal={Advances in neural information processing systems},
  volume={31},
  year={2018}
}

@article{kuaishou2026agentx,
  title={AgentX: Towards Agent-Driven Self-Iteration of Industrial Recommender Systems},
  author={Lao, Changxin and Pan, Fei and Ma, Guozhuang and Li, Han and Lin, Huihuang and Shi, Jijun and Zhao, Kangzhi and Gai, Kun and Zhou, Mo and Zhou, Qinqin and others},
  journal={arXiv preprint arXiv:2606.26859},
  year={2026}
}

@article{zhao2025llmagentrec_survey,
  title={A survey on llm-powered agents for recommender systems},
  author={Peng, Qiyao and Liu, Hongtao and Huang, Hua and Yang, Qing and Shao, Minglai},
  journal={arXiv preprint arXiv:2502.10050},
  year={2025}
}

@inproceedings{wilhelm2018youtubedpp,
  title={Practical diversified recommendations on youtube with determinantal point processes},
  author={Wilhelm, Mark and Ramanathan, Ajith and Bonomo, Alexander and Jain, Sagar and Chi, Ed H and Gillenwater, Jennifer},
  booktitle={Proceedings of the 27th ACM International Conference on Information and Knowledge Management},
  pages={2165--2173},
  year={2018}
}

\end{document}